\PassOptionsToPackage{table}{xcolor}
\documentclass[10pt,twocolumn,letterpaper]{article}

\usepackage[pagenumbers]{cvpr}   % arXiv version

\usepackage{graphicx}
\usepackage{amsmath}
\usepackage{amssymb}
\usepackage{booktabs}
\usepackage{multirow}
\usepackage{makecell}
\usepackage{xspace}
\usepackage{pifont}
\usepackage{xcolor}
\usepackage{colortbl}
\usepackage{tikz}
\usepackage{stfloats} % allow [b] placement for table*/figure*
\usetikzlibrary{arrows.meta,positioning,fit,calc,backgrounds,shapes.misc,patterns}

\newcommand{\name}{EgoManeuver\xspace}
\newcommand{\method}{DenseEgo\xspace}

\newcommand{\cls}[1]{\textit{#1}}

\newcommand{\cmark}{\ding{51}}

\newcommand{\pmstd}[1]{\scriptsize{$\pm$#1}}
\newcommand{\ci}[2]{\scriptsize{[#1, #2]}}
\definecolor{oursrow}{HTML}{E8F1FA} % soft blue highlight for our method (was gray 0.92)

\newcommand{\para}[1]{\noindent\textbf{#1}\hspace{0.4em}}



\newcommand{\nSeg}{86{,}624}
\newcommand{\nSegK}{86.6K}
\newcommand{\nHours}{190.8}
\newcommand{\nTrain}{69{,}219}
\newcommand{\nVal}{5{,}740}
\newcommand{\nTest}{11{,}665}
\newcommand{\hTrain}{149.9}
\newcommand{\hVal}{13.5}
\newcommand{\hTest}{27.4}
\newcommand{\nCountries}{26}

\newcommand{\nTestMoving}{7{,}763}

\newcommand{\hUS}{172.8}

\newcommand{\cntStraight}{45{,}060}

\newcommand{\cntMergeRight}{46}

\newcommand{\cntNudgeRight}{10}

\newcommand{\nKinRemoved}{1{,}694}
\newcommand{\nMotionRemoved}{53}

\newcommand{\oursLat}{82.7}
\newcommand{\oursLong}{83.5}

\newcommand{\gainVmaeLong}{5.0}

\definecolor{cvprblue}{rgb}{0.21,0.49,0.74}
\usepackage[pagebackref,breaklinks,colorlinks,allcolors=cvprblue]{hyperref}

\def\paperID{12874}
\def\confName{CVPR}
\def\confYear{2027}

\title{Fine-Grained Ego-Vehicle Action Recognition from Driving Video}

\author{Zhipeng Bao \quad Wenjie Zhao \quad Qianwen Li$^{*}$\\
University of Georgia, United States\\
{\tt\small zb28097@uga.edu, Cami.Li@uga.edu}\\
{\small $^{*}$Corresponding author}
}

\begin{document}
\twocolumn[{%
\renewcommand\twocolumn[1][]{#1}%
\maketitle
\begin{center}
\centering
\captionsetup{type=figure}
\includegraphics[width=\textwidth]{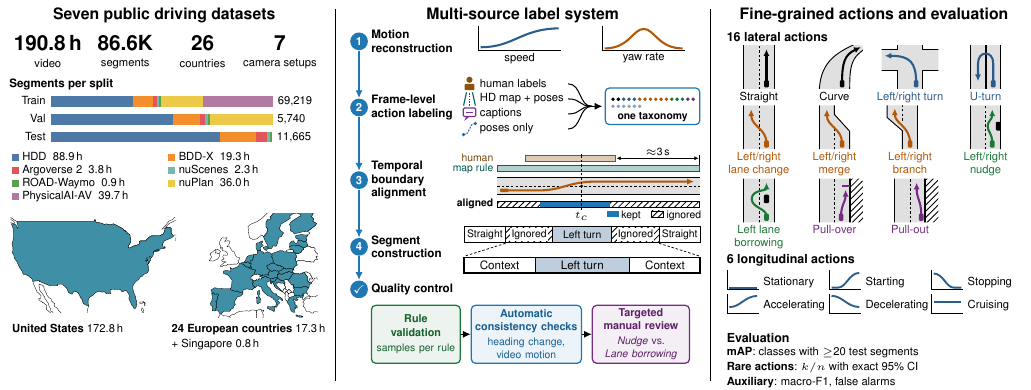}
\vspace{-2pt}
\captionof{figure}{\textbf{Overview of the \name{} benchmark.}
Left: Seven public driving datasets contribute 190.8 hours of video and 86.6K segments across 26 countries, with segment counts shown by split and source.
Middle: A multi-source label system reconstructs ego motion, converts heterogeneous annotations into a unified action taxonomy, aligns temporal boundaries, constructs action segments, and performs quality control.
Right: The benchmark defines 16 lateral and 6 longitudinal action classes, with an evaluation protocol covering common and rare actions.}
\label{fig:overview}
\vspace{6pt}
\end{center}%
}]

\begin{abstract}
Fine-grained ego-vehicle action recognition turns large-scale driving video into structured, searchable driving scenarios, providing a foundation for discovering critical maneuvers and systematically developing and evaluating autonomous driving models. However, large-scale driving videos often lack explicit ego-action labels, and existing datasets vary in action definitions, annotation granularity, and available supervision. We introduce \name, a unified benchmark of 16 lateral and 6 longitudinal actions constructed from seven public driving datasets. Our multi-source labeling system leverages existing annotations and complementary evidence available in each source dataset, including vehicle signals, ego poses, HD-map topology, and captions, to assign labels under a unified fine-grained action taxonomy. We further standardize temporal action boundaries across datasets and validate the resulting labels through kinematic consistency checks and targeted human review. The resulting benchmark contains \nSegK{} segments spanning \nHours{}\,h across \nCountries{} countries and provides a long-tail-aware evaluation protocol.
We also introduce \method, a video-based baseline that combines spatially aware representations with dense per-step action supervision.
On the \name{}, \method achieves \oursLat{} lateral and \oursLong{} longitudinal mAP, outperforming the strongest existing video model by 1.5 and 5.0 points, respectively.
Finally, we demonstrate one way to integrate \method with a language-model agent for retrieving driving videos that match user-specified actions and scenarios.
All code, datasets, and model weights will be publicly released.

\end{abstract}

\section{Introduction}
\label{sec:intro}

Fine-grained ego-vehicle action recognition is important for autonomous driving, both for developing data-driven models and for understanding vehicle interactions with the surrounding environment.
First, recognizing specific maneuvers enables action-based data curation, allowing large collections of driving videos to be organized and selectively retrieved for model training and evaluation.
For example, GenAD~\cite{yang2024genad} uses an action classifier trained on HDD~\cite{ramanishka2018hdd} to annotate web driving videos.
Such annotations are particularly relevant to driving world models, which learn to predict future scenes conditioned on ego actions, as demonstrated by GAIA-1~\cite{hu2023gaia1} and Vista~\cite{gao2024vista}.
Fine-grained action labels can provide more specific supervision for learning how different maneuvers affect future scene evolution.
Second, understanding ego actions can help distinguish changes caused by the ego vehicle from those caused by the surrounding environment, supporting scene interpretation and motion prediction.
For instance, Zhang et al.~\cite{zhang2024decouple} use ego actions to disentangle camera and pedestrian motion for pedestrian trajectory prediction, while FlowAD~\cite{guo2026flowad} incorporates ego-motion feedback into scene modeling to improve perception and planning.
However, fine-grained ego-action labels are often unavailable in large-scale driving recordings, motivating scalable recognition of diverse, fine-grained actions directly from video.

Despite the abundance of public driving datasets, supervision for ego-vehicle actions remains fragmented. Existing datasets differ in both action definitions and annotation granularity. HDD~\cite{ramanishka2018hdd} provides temporally localized annotations for 11 goal-oriented actions, including lane changes, merges, and branches. ROAD~\cite{singh2023road} and ROAD-Waymo~\cite{khan2024roadwaymo} provide frame-level ego-action labels but use broader categories such as moving left or right. Brain4Cars~\cite{jain2015brain4cars}, in contrast, labels short video clips with five maneuver categories for anticipating actions before they occur. Consequently, actions with similar vehicle motion but different relationships to road topology are not consistently distinguished across datasets.
Other datasets provide complementary forms of supervision. BDD-X~\cite{kim2018bddx} describes driving actions in natural-language captions. Other works provide high-level action descriptions~\cite{chi2025impromptu,huang2026nureasoning}. Large-scale datasets such as nuScenes, Argoverse~2, and nuPlan provide ego poses and HD maps from which action labels can be derived~\cite{caesar2020nuscenes,wilson2021argoverse2,karnchanachari2024nuplan}. Although these datasets contain rich evidence of ego-vehicle actions, their heterogeneous representations hinder the construction of a unified, fine-grained action benchmark.

We introduce \name, a benchmark for fine-grained ego-vehicle action recognition. We develop a multi-source labeling system that uses existing annotations and complementary evidence, such as ego-motion measurements and HD-map lane topology, from seven public driving datasets to assign action labels according to a unified fine-grained taxonomy. The system standardizes temporal boundaries across sources and applies kinematic consistency checks and targeted human review to improve label reliability. The resulting benchmark contains 16 lateral and 6 longitudinal action classes, with \nSegK{} segments spanning \nHours{}\,h across \nCountries{} countries.

We further introduce \method, a baseline for \name.
Built on V-JEPA~2~\cite{assran2025vjepa2}, \method combines spatially aware video representations with dense per-step action supervision to capture both road-relative spatial cues and temporal vehicle behavior. On \name, \method achieves \oursLat{} lateral and \oursLong{} longitudinal mAP, matching the strongest existing video model laterally and improving longitudinal mAP by \gainVmaeLong{} points. It further achieves state-of-the-art performance among video-only methods on HDD and ROAD-Waymo, reaching 51.3 and 63.8 mAP, respectively. Our contributions are summarized as follows:
\begin{itemize}
    
\item \name, a unified benchmark that provides standardized fine-grained ego-action annotations and supports model training and evaluation, including the assessment of rare actions.

\item \method, a strong video-based baseline that leverages dense per-step supervision and spatial-temporal representations for fine-grained ego-action recognition, supporting automatic annotation and targeted scenario mining in unlabeled driving videos.

\end{itemize}

\section{Related Work}
\label{sec:related}

\begin{table*}[htbp]
\centering

\caption{\textbf{Comparison of ego-action datasets.}
Lat./Long.: numbers of lateral/longitudinal action classes.
Lane-level, Obstacle, and Curb indicate actions involving lane topology, obstacle avoidance, and roadside stopping or reentry, respectively.}

\label{tab:datasets}
\small
\setlength{\tabcolsep}{4pt}
\begin{tabular}{@{}lllcccccl@{}}
\toprule
Dataset & Size & Coverage & Lat. & Long. &
Lane-level & Obstacle & Curb & Task \\
\midrule
Brain4Cars~\cite{jain2015brain4cars}
& 700 events & 2 US states & 5 & --
& & & & Anticipation \\

HDD~\cite{ramanishka2018hdd}
& 104\,h & SF Bay Area & 8 & --
& \cmark & & & Detection \\

BDD-OIA~\cite{xu2020bddoia}
& 22.9K clips & US & 2 & 2
& & & & Action reasoning \\

ROAD~\cite{singh2023road}
& 2.8\,h & Oxford & 5 & 2
& & & & Detection \\

ROAD-Waymo~\cite{khan2024roadwaymo}
& 5.5\,h & 4 US cities & 4 & 2
& & & & Detection \\

DriveLM-nuScenes~\cite{sima2024drivelm}
& 4.9K frames & 2 cities & 5 bins & 5 bins
& & & & QA \\

Impromptu VLA~\cite{chi2025impromptu}
& 80K clips & 8 datasets & 5 & 4
& & & & QA \\

nuReasoning~\cite{huang2026nureasoning}
& 105\,h & 4 US cities & 7 & 9
& & in-lane & & QA \\
\midrule
\rowcolor{oursrow}
\name (ours)
& 190.8\,h & 26 countries & 16 & 6
& \cmark & \cmark & \cmark & Recognition \\
\bottomrule
\end{tabular}
\end{table*}

\para{Ego-action datasets.}

HDD~\cite{ramanishka2018hdd} distinguishes lane changes, merges, and branches for action detection~\cite{xu2019trn,wang2023mat,cao2023e2eload,cao2026e2eloadpp}, whereas ROAD~\cite{singh2023road} and ROAD-Waymo~\cite{khan2024roadwaymo} group such movements into broader left/right actions.
Brain4Cars~\cite{jain2015brain4cars} and DAAD~\cite{wasi2024daad} focus on action anticipation.
DriveLM~\cite{sima2024drivelm} and LingoQA~\cite{marcu2024lingoqa} incorporate ego-action descriptions and justifications into driving question answering, while nuReasoning~\cite{huang2026nureasoning} and Impromptu VLA~\cite{chi2025impromptu} includes action-related supervision for planning-oriented reasoning.
Together, these works advance ego-action understanding through detection, anticipation, and reasoning.
However, their action definitions, annotation granularity, and temporal localization differ, making consistent recognition and evaluation of fine-grained ego actions across datasets difficult.

\para{Video-based action recognition.}
Video action recognition has been studied using 2D CNNs with temporal aggregation~\cite{wang2016tsn}, 3D CNNs that jointly model spatial and temporal features~\cite{carreira2017i3d,feichtenhofer2019slowfast,feichtenhofer2020x3d}, and video transformers~\cite{bertasius2021timesformer,arnab2021vivit,li2022mvitv2}.
Beyond architectural advances, self-supervised methods such as VideoMAE~\cite{tong2022videomae,wang2023videomaev2} and V-JEPA~\cite{bardes2024vjepa,assran2025vjepa2} learn video representations without action labels.
Recent benchmarks also evaluate vision-language models on ego-action understanding ~\cite{yang2025vsibench,lin2025camerabench,fruhwirthreisinger2025stsbench,xie2025drivebench}.
Despite these advances, fine-grained ego-action recognition remains insufficiently evaluated, particularly for actions that require understanding road topology beyond vehicle motion alone.

\section{The \name Benchmark}
\label{sec:benchmark}

\begin{figure*}[t]
\centering
\includegraphics[width=\textwidth]{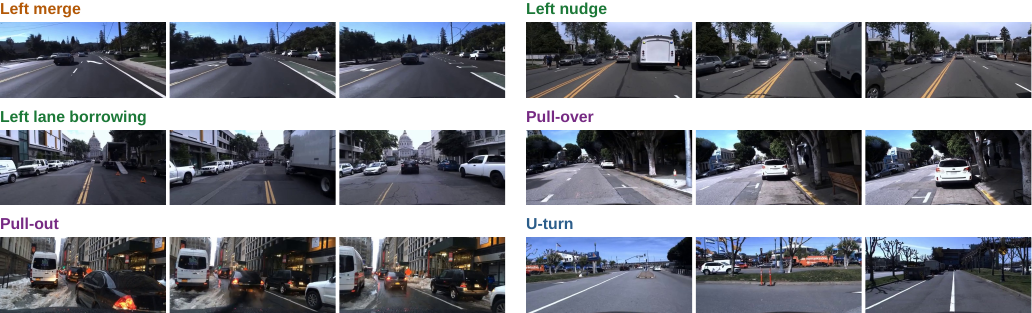}
\caption{\textbf{Examples of fine-grained ego-vehicle actions from EgoManeuver.}}
\label{fig:examples}
\end{figure*}

%-------------------------------------------------------------------------

\subsection{Task and Taxonomy}
\label{sec:task}

\para{Task.}
Fine-grained ego-action recognition characterizes ego-vehicle behavior along two dimensions: lateral and longitudinal motion. From a video segment $x$, the model independently estimates the lateral and longitudinal actions as probability distributions $\hat{\mathbf{p}}^{\mathrm{lat}}(x)$ over $\mathcal{C}^{\mathrm{lat}}$ and $\hat{\mathbf{p}}^{\mathrm{lon}}(x)$ over $\mathcal{C}^{\mathrm{lon}}$. For stationary segments, the model is not required to predict a lateral action.

\para{Taxonomy.}
The lateral taxonomy consists of 16 action classes (\cref{fig:overview}): \cls{Lane keeping} (\cls{straight}, \cls{curve}), \cls{Left/right turn}, \cls{U-turn}, \cls{Left/right lane change}, \cls{Left/right merge}, \cls{Left/right branch}, \cls{Left/right nudge}, \cls{Left lane borrowing}, \cls{Pull-over}, and \cls{Pull-out}. These classes are defined jointly by ego motion and road topology. A \cls{Lane change} denotes a transition to an adjacent continuing lane; a \cls{Merge} denotes a transition from an ending lane or onto another road; and a \cls{Branch} denotes entry into a lane that diverges from the current path, such as a turn pocket or exit. A \cls{Nudge} is a lateral displacement within the current lane for obstacle avoidance, whereas \cls{Lane borrowing} is the temporary use of an adjacent or oncoming lane followed by a return to the original lane. \cls{Pull-over} and \cls{Pull-out} denote transitions from traffic to a roadside stop and back into traffic, respectively (\cref{fig:examples}).

The six longitudinal action classes are \cls{Stationary}, \cls{Starting}, \cls{Stopping}, \cls{Accelerating}, \cls{Decelerating}, and \cls{Cruising}. \cls{Stationary} denotes a stopped state, while \cls{Starting} and \cls{Stopping} represent transitions from stationary to moving and from moving to stationary, respectively. \cls{Accelerating} and \cls{Decelerating} denote increases and decreases in speed, respectively, whereas \cls{Cruising} represents motion without substantial speed change. Following prior work on kinematics-based driving action labeling~\cite{degelder2020scenario,ma2022driving}, we assign longitudinal action labels based on ego speed and its change over each segment. Let $v(t)$ denote ego speed, and $\bar v_0$ and $\bar v_1$ the mean speeds over the first and last second of a segment, respectively. We set the stationary threshold to $v_{\mathrm{s}}=0.3$\,m/s and the speed-change threshold to $\Delta=10$\,km/h. We assign the label as

\begin{equation}
y^{\mathrm{lon}} =
\begin{cases}
\textit{Stationary} & \text{if } \max_t v(t) < v_{\mathrm{s}},\\
\textit{Starting} & \text{if } \bar v_0 < v_{\mathrm{s}} \text{ and } \bar v_1 \ge \Delta/2,\\
\textit{Stopping} & \text{if } \bar v_0 \ge \Delta/2 \text{ and } \bar v_1 < v_{\mathrm{s}},\\
\textit{Accelerating} & \text{if } \bar v_1 - \bar v_0 \ge \Delta,\\
\textit{Decelerating} & \text{if } \bar v_0 - \bar v_1 \ge \Delta,\\
\textit{Cruising} & \text{otherwise},
\end{cases}
\label{eq:longitudinal}
\end{equation}

\subsection{Evidence-Guided Action Labeling}
\label{sec:labels}
The seven datasets shown in \cref{fig:overview} provide different types of evidence for ego-action labeling, including human annotations, ego-motion data, HD-map lane topology, and natural-language captions. We use this evidence in a four-step pipeline to generate action labels.

\para{Ego kinematics extraction.}
We reconstruct ego speed, acceleration, and yaw rate from available ego poses, CAN-bus signals, or GPS/gyroscope measurements. The reconstructed motion data are aligned with the video at 10\,Hz to support action labeling and kinematic consistency checks.

\para{Frame-level action labeling.}
We assign lateral and longitudinal action labels at 10\,Hz. Longitudinal labels are derived from speed and acceleration rules (Eq. \ref{eq:longitudinal}). For lateral actions, we use HD-map lane topology in Argoverse 2, nuScenes, and nuPlan. We match ego trajectories to lane graphs using a hidden Markov model with Viterbi decoding~\cite{newson2009hidden}:
\begin{equation}
\ell^{*}_{1:T}
=
\arg\min_{\ell_{1:T}}
\left[
\sum_{t=1}^{T} e_t(\ell_t)
+
\sum_{t=2}^{T} c(\ell_{t-1},\ell_t)
\right],
\label{eq:hmm}
\end{equation}
where $\ell_t$ is the assigned lane at time $t$, $e_t$ penalizes lateral and heading offsets from lane $\ell_t$, and $c$ penalizes transitions based on lane connectivity. The matched trajectory and lane topology distinguish \cls{Lane keeping}, \cls{Left/right turn}, \cls{U-turn}, \cls{Left/right lane change}, \cls{Left/right merge}, \cls{Left/right branch}, \cls{Pull-over}, and \cls{Pull-out}.

For caption-based labeling, we match time-stamped BDD-X captions to predefined action patterns, such as "changes into the left lane" for lane changes, and assign the corresponding labels to the caption intervals. For pose-only labeling, we use ego heading change, yaw rate, turning radius, and trajectory curvature to identify lane keeping, turns, and U-turns. Lane-level maneuvers are ignored because ego poses alone cannot determine lane transitions.

\para{Temporal boundary alignment.}
The action labels obtained in the previous step may have inconsistent temporal boundaries because they are derived from different annotation evidence, with discrepancies of up to several seconds (\cref{fig:overview}). To reduce this boundary uncertainty, we retain only the core interval of each lateral action for training and mark the frames outside the core interval as ignored.

For map-derived lane changes, the core interval is defined around the inferred lane-crossing time $t_c$. For turns with reliable heading measurements, the core interval covers the central 80\% of the total heading change within the original event $[t_s,t_e]$:
\begin{equation}
\begin{aligned}
\mathcal{T}_{\mathrm{lc}}
&= [t_c - 1.5\,\mathrm{s},\,t_c + 1\,\mathrm{s}]
\cap [t_s,t_e],\\
\mathcal{T}_{\mathrm{turn}}
&= \left\{t \in [t_s,t_e]:
0.1 \le
\frac{\psi(t)-\psi(t_s)}
{\psi(t_e)-\psi(t_s)}
\le 0.9
\right\},
\end{aligned}
\label{eq:boundaries}
\end{equation}
where $\psi(t)$ denotes the ego-vehicle heading. Only core intervals lasting at least 0.5\,s are retained.

\para{Segment construction.}
Using the aligned frame-level labels, we construct video segments for benchmark training and evaluation. Each non-lane-keeping lateral action defines a segment, extended by up to 1\,s of context on either side without overlapping neighboring actions. Lane-keeping intervals are divided into 3--10\,s segments. Each resulting segment is then assigned a longitudinal label based on its speed profile using \cref{eq:longitudinal}.

%-------------------------------------------------------------------------
\subsection{Quality Control}
\label{sec:quality}

We validate the generated action labels through automatic consistency checks and targeted manual review.

\textbf{Automatic consistency checks.}
We first verify that the observed heading change is consistent with the assigned lateral action. For each segment spanning $[t_0,t_1]$, we compute
\begin{equation}
\Delta\psi = \int_{t_0}^{t_1} \big(\omega(t)-b\big)\,dt,
\label{eq:heading}
\end{equation}
where $\omega(t)$ denotes yaw rate and $b$ is the bias estimated from stationary frames. 
\cls{Left/right turn} and \cls{U-turn} require large heading changes in the corresponding direction, whereas other actions, such as \cls{Lane keeping (straight)} and \cls{Lane change}, require smaller heading changes.
We remove segments violating these constraints. These checks remove 1747 kinematically inconsistent segments.

\textbf{Targeted manual review.}
We manually review actions that cannot be reliably distinguished from ego-motion data alone when HD maps are unavailable. For example, HDD avoidance annotations provide candidates for \cls{Nudge} and \cls{Lane borrowing}, which exhibit similar ego-motion patterns. We therefore inspect the corresponding videos to correct or exclude ambiguous labels.

% Finally, we perform a stratified manual audit of the test set, covering all rare and obstacle-related actions together with randomly sampled examples from every remaining class.
% Among 272 audited segments, all longitudinal labels and 268 of the 271 decidable lateral labels were correct, with at least 93\% precision for every lateral class.
% The identified errors were corrected and the single undecidable segment was removed before evaluation.

%-------------------------------------------------------------------------
\subsection{Splits and Statistics}
\label{sec:stats}

% The test split consists of the official validation and test splits of the sources (for HDD, its test sessions); the validation split holds a fixed 10\% of the training drives of every source except PAI.
% nuPlan contributes only training and validation data, and PAI, as its license requires, only training data.

\begin{figure*}[t]
\centering
\includegraphics[width=\textwidth]{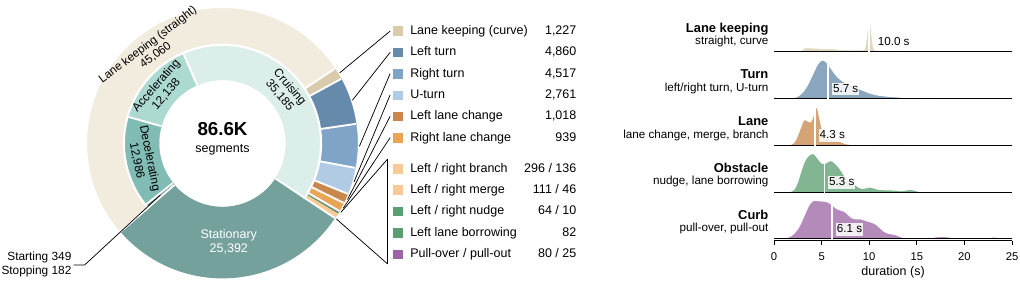}
\caption{\textbf{Action distribution and duration statistics of \name.}
Left: Longitudinal and lateral action distributions; set-valued labels represent multiple valid candidate classes.
Right: Duration distributions by action family, with median values annotated.}
\label{fig:stats}
\end{figure*}

\name contains \nSeg{} segments (\nHours{}\,h) from seven camera setups across \nCountries{} countries, including \nTrain{} training (\hTrain{}\,h), \nVal{} validation (\hVal{}\,h), and \nTest{} test segments (\hTest{}\,h).
The action distribution is highly imbalanced (\cref{fig:stats}): \cls{Lane keeping (straight)} accounts for \cntStraight{} segments, compared with only \cntMergeRight{} for \cls{Right merge} and \cntNudgeRight{} for \cls{Right nudge}.
Action durations also vary by family, with \cls{Lane keeping} segments lasting longer than most maneuver segments.

\subsection{Action Recognition Evaluation}
\label{sec:protocol}

We evaluate lateral and longitudinal actions separately. Lateral evaluation uses only the moving test segments because lateral actions are undefined for stationary samples. Longitudinal evaluation uses all test segments.

\para{Primary metrics.}
We use mean average precision (mAP) to evaluate action classes with at least 20 positive test segments. For each class $c$, average precision $\mathrm{AP}_c$ is computed from the precision--recall curve by ranking segments according to their predicted probability $\hat p_c$~\cite{everingham2010pascal}. We calculate mAP as
\begin{equation}
\mathrm{mAP}
=
\frac{1}{|\mathcal{H}|}
\sum_{c\in\mathcal{H}} \mathrm{AP}_c,
\qquad
\mathcal{H}
=
\{c \mid n_c \ge 20\},
\label{eq:map}
\end{equation}
where $n_c$ is the number of positive test segments for class $c$, $\mathcal{H}$ is the set of evaluated classes.
% , and $\mathcal{B}$ contains the background class. We designate \cls{Lane keeping (straight)} and \cls{Stationary} as background classes. These frequent, easily recognized classes are excluded from mAP to prevent their high AP from inflating the overall metric and obscuring performance on other actions.

\para{Low-sample actions.}
The seven lateral classes with fewer than 20 positive test segments are \cls{U-turn}, \cls{Left/right merge}, \cls{Left/right nudge}, \cls{Pull-over}, and \cls{Pull-out}. Because AP estimates are unstable with so few examples, we instead report the number of correct top-1 predictions $k$ out of $n$ test segments for each class, together with exact 95\% Clopper--Pearson confidence intervals~\cite{clopper1934exact}.

\para{Auxiliary metrics.}
We additionally report macro-F1 over the classes. Since \cls{Lane keeping (straight)} is the most common lateral action, incorrectly identifying these segments can lead to numerous false detections. We therefore report the false-alarm rate as the fraction of \cls{Lane keeping (straight)} segments predicted as non-lane-keeping actions.

To evaluate whether models can recognize the general type of a maneuver even when they misclassify its fine-grained action, we group related lateral classes into four action families: Turn, Lane, Obstacle, and Curb.
Each family contains the corresponding fine-grained classes (Fig. \ref{fig:stats}). For a family $F$, we sum the predicted probabilities of its member classes, $s_F=\sum_{c\in F}\hat p_c$, and compute family-level AP by treating any member class as positive.

% All results are averaged over three random seeds. We estimate 95\% confidence intervals using 1,000 bootstrap resamples of complete drives within each data source, with paired resampling for method comparisons.

\section{\method: Spatially Aware Readout and Dense Action Supervision}
\label{sec:baseline}

\begin{figure}[t]
\centering
\includegraphics{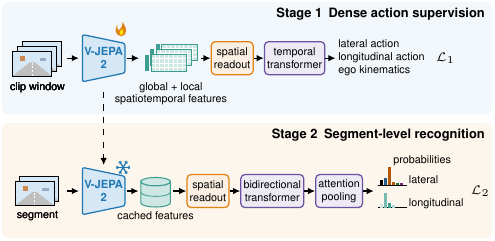}
\caption{\textbf{Architecture and two-stage training of \method.}
Stage 1 fine-tunes V-JEPA~2 using dense per-step action and ego-kinematic supervision.
Stage 2 freezes the adapted encoder and trains a segment-level recognition head with spatial readout, temporal modeling, and attention pooling.}
\label{fig:baseline}
\end{figure}

\method builds on the pretrained V-JEPA~2 video encoder~\cite{assran2025vjepa2} to recognize fine-grained ego actions from video.
It combines a spatially aware readout with two-stage training (\cref{fig:baseline}).
Stage~1 adapts the encoder using dense frame-level action supervision, while Stage~2 trains a segment-level action recognition head.

% Table 2 (tab:main) is defined here so that it floats to the bottom of page 6
\begin{table*}[b]
\centering
\caption{\textbf{Comparison on the \name{} test split} (11{,}665 segments).
Lateral and longitudinal mAP are computed over eight and five non-background classes, respectively.
Results are reported as mean $\pm$ standard deviation over three seeds, with drive-level bootstrap 95\% confidence intervals.
F1 denotes macro-F1.
Family AP evaluates four lateral action families: Turn (left/right turn, U-turn), Lane (lane change, merge, branch), Obst. (nudge, lane borrowing), and Curb (pull-over, pull-out).
FA is the false-alarm rate on \cls{Straight}, and Rare is the mean accuracy over seven low-sample lateral classes.
IN-1K: ImageNet-1K; K400: Kinetics-400; SSL: self-supervised V-JEPA~2.
Best results are highlighted in bold red.}
\label{tab:main}
\small
\setlength{\tabcolsep}{2.6pt}
\begin{tabular}{@{}ll l@{\hspace{2pt}}l l@{\hspace{2pt}}l cc cccc cc@{}}
\toprule
 & & \multicolumn{2}{c}{Lateral} & \multicolumn{2}{c}{Longitudinal} & \multicolumn{2}{c}{F1} & \multicolumn{4}{c}{Family AP} & & \\
\cmidrule(lr){3-4}
\cmidrule(lr){5-6}
\cmidrule(lr){7-8}
\cmidrule(lr){9-12}
Method & Pre-train & mAP$\uparrow$ & 95\% CI & mAP$\uparrow$ & 95\% CI & Lat.$\uparrow$ & Long.$\uparrow$ & Turn$\uparrow$ & Lane$\uparrow$ & Obst.$\uparrow$ & Curb$\uparrow$ & FA$\downarrow$ & Rare$\uparrow$ \\
\midrule
TSN-R50~\cite{wang2016tsn} & IN-1K & 48.9\pmstd{0.6} & \ci{45.6}{52.9} & 31.9\pmstd{0.7} & \ci{30.5}{33.5} & 51.6 & 41.0 & 93.8 & 66.7 & 28.0 & 4.7 & 6.3 & 17.7 \\
I3D-R50~\cite{carreira2017i3d} & K400 & 72.9\pmstd{0.8} & \ci{69.5}{76.5} & 65.6\pmstd{0.8} & \ci{61.3}{70.4} & 69.2 & 66.6 & 98.5 & 86.6 & 62.7 & 25.5 & 2.3 & 35.1 \\
X3D-M~\cite{feichtenhofer2020x3d} & K400 & 68.7\pmstd{0.7} & \ci{65.1}{71.9} & 67.8\pmstd{0.8} & \ci{63.4}{72.6} & 66.9 & 70.6 & 99.2 & 82.6 & 53.9 & 28.9 & 3.0 & 43.1 \\
SlowFast-R50~\cite{feichtenhofer2019slowfast} & K400 & 77.1\pmstd{1.3} & \ci{73.5}{80.2} & 76.9\pmstd{0.4} & \ci{72.4}{82.0} & 72.8 & 76.0 & 99.6 & 87.6 & 59.0 & 50.4 & 1.2 & 39.6 \\
TimeSformer-B~\cite{bertasius2021timesformer} & K400 & 72.3\pmstd{1.3} & \ci{68.7}{75.6} & 46.9\pmstd{1.8} & \ci{44.6}{50.4} & 69.0 & 51.2 & 99.1 & 86.5 & 46.1 & 38.3 & 1.8 & 28.7 \\
VideoMAE-B~\cite{tong2022videomae} & K400 & 81.2\pmstd{0.8} & \ci{77.5}{84.7} & 78.5\pmstd{1.2} & \ci{73.8}{83.3} & 76.0 & 78.3 & 99.7 & \textcolor{red}{\textbf{92.5}} & 71.6 & \textcolor{red}{\textbf{59.4}} & 1.1 & 44.9 \\
\midrule
V-JEPA~2-L, frozen, pooled & SSL & 76.2\pmstd{0.6} & \ci{71.0}{80.3} & 81.4\pmstd{1.0} & \ci{77.0}{86.2} & 76.3 & 80.6 & 99.6 & 87.1 & \textcolor{red}{\textbf{79.2}} & 38.2 & 1.4 & \textcolor{red}{\textbf{45.2}} \\
V-JEPA~2-L, frozen, grid & SSL & 77.7\pmstd{1.3} & \ci{72.6}{81.8} & 81.0\pmstd{0.5} & \ci{76.8}{85.9} & 75.7 & 80.2 & 99.6 & 88.1 & 76.3 & 38.0 & 1.1 & 42.2 \\
\rowcolor{oursrow}
\method & SSL & \textcolor{red}{\textbf{82.7}}\pmstd{0.5} & \ci{78.2}{86.2} & \textcolor{red}{\textbf{83.5}}\pmstd{0.5} & \ci{79.1}{88.4} & \textcolor{red}{\textbf{81.2}} & \textcolor{red}{\textbf{82.6}} & \textcolor{red}{\textbf{99.8}} & 91.5 & 78.7 & 54.1 & \textcolor{red}{\textbf{0.9}} & 44.2 \\
\bottomrule
\end{tabular}
\end{table*}

\para{Spatially aware readout.}
Fine-grained lateral actions often depend on road layout as well as ego motion.
To retain spatial cues, we extract a global feature and a grid of spatially pooled features from the encoder at each temporal step, rather than relying on global pooling alone.
A learned attention query aggregates these features into a step representation:
\begin{equation}
\mathbf{h}_t =
\mathrm{Attn}\left(
\mathbf{q},
\{\mathbf{W}\mathbf{z}_{t,k}+\mathbf{e}_k\}_{k=1}^{K}
\right),
\label{eq:readout}
\end{equation}
where $\mathbf{z}_{t,k}$ is the $k$-th of $K$ pooled visual features at step $t$, $\mathbf{W}$ is a projection matrix, $\mathbf{e}_k$ encodes its spatial position, and $\mathbf{q}$ is a learned query. The output $\mathbf{h}_t$ represents the visual information at step $t$.

\para{Stage 1: Dense action supervision.}
We use frame-level action labels to fine-tune the encoder for frame-wise action recognition.
A temporal transformer processes the frame-level representations and predicts lateral and longitudinal actions, with auxiliary supervision from ego kinematics and HDD goal-oriented action labels where available.
% To accommodate set-valued action labels, we use
% \begin{equation}
% \mathcal{L}_{\mathrm{set}}(\hat{\mathbf{p}},S)
% =
% -\log\sum_{c\in S}\hat p_c,
% \label{eq:setloss}
% \end{equation}
% where $S$ is the set of valid classes and $\hat p_c$ is the predicted probability of class $c$.
% For an exact label $S=\{y\}$, the loss reduces to cross-entropy.
The Stage~1 objective combines action classification and auxiliary supervision:
\begin{equation}
\mathcal{L}_1 =
\mathcal{L}_{\mathrm{ce}}^{\mathrm{lat}}
+\mathcal{L}_{\mathrm{ce}}^{\mathrm{lon}}
+\mathcal{L}_{\mathrm{ce}}^{\mathrm{goal}}
+\lambda\mathcal{L}_{\mathrm{kin}},
\label{eq:stage1}
\end{equation}
where $\mathcal{L}_{\mathrm{ce}}^{\mathrm{lat}}$, $\mathcal{L}_{\mathrm{ce}}^{\mathrm{lon}}$, and $\mathcal{L}_{\mathrm{ce}}^{\mathrm{goal}}$ are cross-entropy losses for lateral action, longitudinal action, and HDD goal-oriented action classification, respectively. $\mathcal{L}_{\mathrm{kin}}$ is the smooth-$L_1$ loss for kinematic regression, with $\lambda=0.5$.
% Each term is averaged over steps with valid targets, with goal-oriented supervision applied only to HDD.
We first train the heads with a frozen encoder, then fine-tune the full model.

\para{Stage 2: Segment-level recognition.}
We freeze the adapted encoder and cache its visual features.
For each segment, the spatial readout produces step representations, which a bidirectional temporal transformer integrates across time.
Attention pooling yields a segment representation for separate lateral and longitudinal classification heads.
We train both heads using the cross-entropy loss $\mathcal{L}_2=\mathcal{L}_{\mathrm{ce}}^{\mathrm{lat}}+\mathcal{L}_{\mathrm{ce}}^{\mathrm{lon}}$.

\begin{table*}[t]
\centering
\caption{\textbf{Per-class AP on the \name{} test split.}
Results are averaged over three seeds; $n$ denotes test samples.
L/R: left/right; T: turn; LC: lane change; Br: branch; Bor: lane borrowing; Mg: merge; Ng: nudge; PO/POut: pull-over/pull-out.
$^\dagger$Background classes and $^\ast$rare classes ($n<20$) are excluded from mAP.
Best results are highlighted in bold red.}
\label{tab:perclass}
\setlength{\tabcolsep}{2.5pt}
\renewcommand{\arraystretch}{1.15}
\resizebox{\textwidth}{!}{%
\begin{tabular}{@{}l*{22}{c}@{}}
\toprule
 & \multicolumn{9}{c}{Lateral ($n\geq20$)}
 & \multicolumn{7}{c}{Lateral ($n<20$)}
 & \multicolumn{6}{c}{Longitudinal} \\
\cmidrule(lr){2-10}
\cmidrule(lr){11-17}
\cmidrule(lr){18-23}
Method
& Str$^\dagger$ & Cur & L-T & R-T & L-LC & R-LC & L-Br & R-Br & L-Bor
& UT & L-Mg & R-Mg & L-Ng & R-Ng & PO & POut
& Stat$^\dagger$ & Start & Stop & Acc & Dec & Cru \\
\midrule
$n$
& 6382 & 100 & 412 & 462 & 123 & 117 & 44 & 32 & 25
& 18 & 17 & 8 & 11 & 1 & 5 & 6
& 3902 & 65 & 20 & 1575 & 1798 & 4241 \\
\midrule
TSN-R50
& 98.7 & 49.0 & 79.2 & 88.2 & 39.8 & 38.2 & 70.5 & 13.3 & 13.2
& 9.8 & 6.8 & 38.4 & 12.1 & 0.1 & 6.4 & 4.2
& 99.0 & 4.8 & 1.5 & 39.3 & 43.2 & 70.6 \\
I3D-R50
& 99.6 & 65.2 & 96.5 & 98.3 & 87.5 & 81.3 & 85.3 & 40.7 & 28.2
& 54.8 & 38.9 & 37.4 & 19.6 & \textcolor{red}{\textbf{1.2}} & 46.8 & 18.4
& 99.7 & 42.6 & 31.6 & 81.5 & 86.2 & 86.1 \\
X3D-M
& 99.7 & 65.8 & 98.2 & 99.3 & 75.1 & 68.5 & 78.4 & 33.6 & 30.6
& 75.9 & 19.4 & 32.5 & 13.2 & 0.2 & 45.1 & 31.4
& 99.8 & 55.8 & 27.7 & 81.3 & 87.6 & 86.4 \\
SlowFast-R50
& 99.7 & 72.9 & 99.2 & 99.4 & 88.1 & 79.3 & 85.4 & 49.5 & 42.7
& 77.3 & 28.7 & 35.5 & 15.2 & 1.1 & 76.4 & 34.3
& 99.9 & 76.0 & 35.6 & 89.7 & 91.7 & 91.6 \\
TimeSformer-B
& 99.6 & 60.3 & 97.1 & 98.5 & 85.2 & 77.7 & 85.6 & 49.3 & 25.1
& 66.3 & 38.0 & 38.6 & 13.6 & 0.4 & 54.4 & 34.7
& 99.9 & 16.5 & 10.3 & 58.2 & 69.4 & 80.1 \\
VideoMAE-B
& 99.8 & 77.3 & 99.4 & 99.7 & \textcolor{red}{\textbf{94.6}} & 86.3 & 89.7 & \textcolor{red}{\textbf{57.5}} & 43.4
& 88.9 & \textcolor{red}{\textbf{55.7}} & \textcolor{red}{\textbf{43.7}} & 22.1 & 0.4 & \textcolor{red}{\textbf{98.9}} & 30.0
& 99.9 & 71.1 & 45.5 & 90.1 & 92.8 & 92.5 \\
\midrule
V-JEPA~2-L, frozen, pooled
& 99.7 & 76.4 & 99.5 & 99.4 & 85.7 & 78.4 & 85.3 & 33.9 & 51.1
& 86.5 & 26.2 & 31.3 & \textcolor{red}{\textbf{45.3}} & 0.2 & 65.8 & 23.1
& 99.9 & 78.5 & 41.5 & 94.5 & 96.1 & 96.1 \\
V-JEPA~2-L, frozen, grid
& 99.8 & 77.2 & 99.5 & 99.5 & 89.0 & 80.0 & 86.2 & 37.1 & 53.4
& 88.6 & 33.0 & 35.1 & 42.2 & 0.2 & 62.5 & 24.2
& \textcolor{red}{\textbf{100.0}} & 79.7 & 38.3 & 94.8 & 96.0 & 96.1 \\
\rowcolor{oursrow}
\method
& \textcolor{red}{\textbf{99.9}} & \textcolor{red}{\textbf{82.2}} & \textcolor{red}{\textbf{99.6}} & \textcolor{red}{\textbf{99.7}} & 91.4 & \textcolor{red}{\textbf{88.0}} & \textcolor{red}{\textbf{93.3}} & 48.8 & \textcolor{red}{\textbf{60.2}}
& \textcolor{red}{\textbf{89.6}} & 44.9 & 38.1 & 32.0 & 0.3 & 81.8 & \textcolor{red}{\textbf{35.9}}
& 99.9 & \textcolor{red}{\textbf{80.0}} & \textcolor{red}{\textbf{45.9}} & \textcolor{red}{\textbf{96.7}} & \textcolor{red}{\textbf{97.8}} & \textcolor{red}{\textbf{97.5}} \\
\bottomrule
\end{tabular}%
}
\end{table*}

\section{Experiments}
\para{Experimental setup.}
We train and evaluate \method and six representative video action recognition models on \name: TSN~\cite{wang2016tsn}, the 3D CNNs I3D~\cite{carreira2017i3d}, SlowFast~\cite{feichtenhofer2019slowfast}, and X3D~\cite{feichtenhofer2020x3d}, and the video transformers TimeSformer~\cite{bertasius2021timesformer} and VideoMAE~\cite{tong2022videomae}.
The comparison models are fine-tuned from public Kinetics-400 checkpoints (ImageNet for TSN), using uniformly sampled frames and class-balanced sampling.
We further conduct ablation studies of \method, evaluate its performance on HDD and ROAD-Waymo, analyze fine-grained action recognition errors, and demonstrate its application to natural-language video retrieval.
Implementation details are provided in the supplementary material.

%-------------------------------------------------------------------------

\subsection{Performance on \name}
\label{sec:main}

\para{Comparison with existing models.}
The six representative video models exhibit considerable performance differences on \name{} (\cref{tab:main,tab:perclass}). VideoMAE achieves the highest lateral and longitudinal mAP among them (81.2 and 78.5), whereas TimeSformer reaches 72.3 lateral mAP but only 46.9 longitudinal mAP, indicating that performance on the two action dimensions does not necessarily align. At the family level, Turn AP is consistently high (93.8--99.7), while Obstacle and Curb AP are considerably lower. Per-class results further reveal difficulties in recognizing lane-level maneuvers, particularly \cls{Right branch}, suggesting that distinguishing actions based on their relationship to road topology and surrounding objects remains challenging.

\para{\method performance.}
\method achieves 82.7 lateral mAP and 83.5 longitudinal mAP, exceeding VideoMAE by 1.5 and 5.0 points, respectively. \method also achieves higher macro-F1 for lateral (81.2 vs. 76.0) and longitudinal actions (82.6 vs. 78.3), while reducing the false-alarm rate on \cls{Straight} (0.9 vs. 1.1). At the class level, \method achieves higher AP for \cls{Curve} (82.2 vs. 77.3), \cls{Left lane borrowing} (60.2 vs. 43.4), \cls{Accelerating} (96.7 vs. 90.1), and \cls{Decelerating} (97.8 vs. 92.8). However, it performs worse than VideoMAE on \cls{Left lane change} and \cls{Right branch}, indicating that its gains do not extend to all lane-level actions.

\subsection{Ablation Studies}
\label{sec:ablation}

\begin{table}[t]
\centering
\caption{\textbf{Ablation study of \method.}
Each cell reports validation / test results, averaged over three seeds.
Rare Acc. denotes mean rare-class accuracy.}
\label{tab:ablation}
\small
\setlength{\tabcolsep}{4pt}
\renewcommand{\arraystretch}{1.12}
\begin{tabular}{@{}lccc@{}}
\toprule
Configuration
& \shortstack{Lat. mAP$\uparrow$\\Val. / Test}
& \shortstack{Long. mAP$\uparrow$\\Val. / Test}
& \shortstack{Rare Acc.$\uparrow$\\Val. / Test} \\
\midrule
\rowcolor{oursrow}
\method (full)
& 92.5 / 82.7
& 95.9 / 83.5
& 40.3 / 44.2 \\
Frozen encoder
& 88.8 / 77.7
& 93.6 / 81.0
& 31.0 / 42.2 \\
Segment labels only
& 90.8 / 81.3
& 94.1 / 82.9
& 31.6 / 43.7 \\
No balanced sampling
& 91.8 / 85.9
& 95.5 / 86.0
& 28.2 / 41.5 \\
\bottomrule
\end{tabular}
\end{table}

\para{Encoder adaptation and dense supervision.}
Fine-tuning the encoder improves test lateral and longitudinal mAP by 5.0 and 2.5 points over the frozen encoder (\cref{tab:ablation}), indicating the benefit of adapting pretrained video representations to ego-action recognition. Compared with segment-level supervision alone, dense per-step supervision yields modest improvements in overall mAP but increases validation rare-class accuracy from 31.6 to 40.3. However, the test gain is smaller (43.7 to 44.2), suggesting that this benefit does not consistently generalize across splits.

% \para{Unified supervision.}
% Unified training uses a shared action taxonomy across all datasets, whereas source-specific federated training supervises only the classes observable in each dataset. Although validation mAP is similar, the federated loss reduces test lateral mAP from 82.7 to 77.1 (\cref{tab:ablation}), while longitudinal mAP remains comparable (83.5 vs. 83.8). This suggests that cross-source negative supervision helps distinguish lateral actions and improves generalization across datasets.

\para{Class balancing.}
Removing class-balanced sampling increases test lateral and longitudinal mAP from 82.7/83.5 to 85.9/86.0, but reduces rare-class accuracy from 40.3 to 28.2 on validation and from 44.2 to 41.5 on test. We therefore retain it to better recognize underrepresented actions, at a small cost in overall mAP.

\begin{figure}[t]
\centering
\includegraphics[width=0.92\columnwidth]{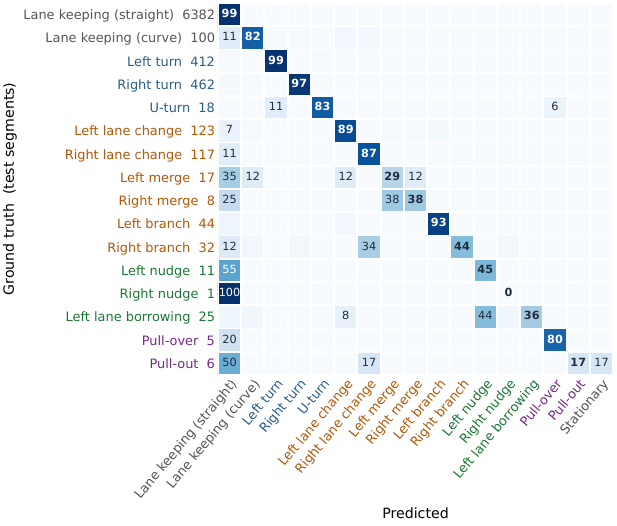}
\caption{\textbf{Confusion matrix of \method on the test split.}
Rows are normalized by ground-truth class, with values shown as percentages.
Numbers beside class names indicate test sample counts.
Confusion is concentrated among several fine-grained lateral actions.}
\label{fig:confusion}
\end{figure}

\begin{figure*}[t]
\centering
\includegraphics[width=0.7\textwidth]{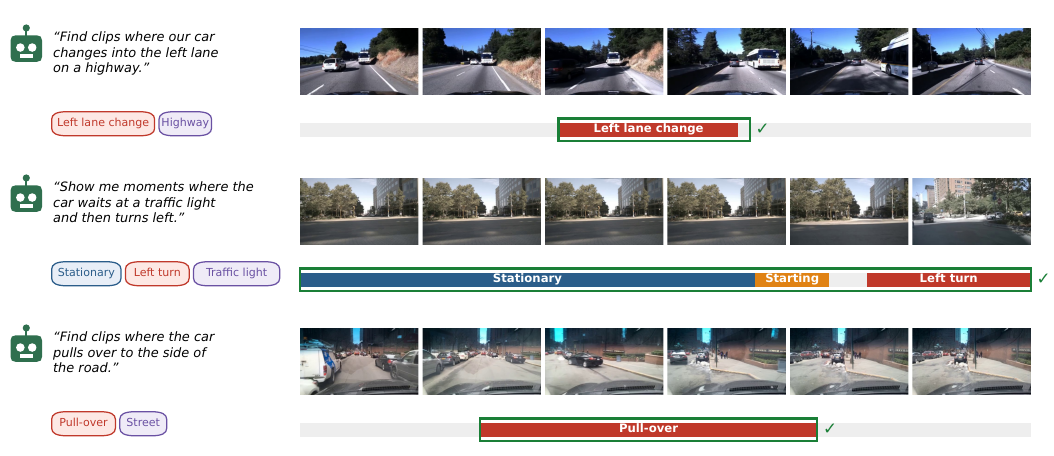}
\caption{\textbf{Natural-language video retrieval agent powered by \method.}
The agent interprets user requests and retrieves matching temporal intervals using \method.
Colored bars show predicted actions.}
\label{fig:agent_retrieval}
\end{figure*}

\subsection{Performance on Existing Benchmarks}
\label{sec:transfer}

\begin{table}[t]
\centering
\caption{\textbf{Comparison on existing benchmarks.}}
\label{tab:transfer}
\small
\setlength{\tabcolsep}{3pt}
\renewcommand{\arraystretch}{0.92}
\begin{tabular}{@{}lcc@{}}
\toprule
& \multicolumn{2}{c}{Frame-level mAP$\uparrow$} \\
\cmidrule(lr){2-3}
Method & HDD & ROAD-Waymo \\
\midrule
E2E-LOAD~\cite{cao2023e2eload} & 48.1 & -- \\
E2E-LOAD++~\cite{cao2026e2eloadpp} & 48.4 & -- \\
SDL~\cite{wang2025sdl} & 49.1 & -- \\
SlowFast~\cite{feichtenhofer2019slowfast} & -- & 52.7 \\
\midrule
\method (mapped) & -- & 56.6 \\
\rowcolor{oursrow}
\method (new head) & \textcolor{red}{\textbf{51.3}} & \textcolor{red}{\textbf{63.8}} \\
\bottomrule
\end{tabular}

\vspace{2pt}
\begin{minipage}{\linewidth}
\footnotesize

\textit{Note:} \textit{Mapped}: the \name{}-trained model's action probabilities are mapped to ROAD-Waymo classes without additional training. \textit{New head}: the \name{}-trained encoder is frozen, and a task-specific classification head is trained on each dataset.

\end{minipage}
\end{table}

We evaluate \method on HDD~\cite{ramanishka2018hdd} and ROAD-Waymo~\cite{khan2024roadwaymo} using their original action labels and evaluation protocols (\cref{tab:transfer}). HDD defines 11 goal-oriented actions, including lane changes, merges, and branches, whereas ROAD-Waymo defines six broader actions, including moving, stopping, left/right turns, and left/right lateral movements, without distinguishing lane changes, merges, or other lane-level maneuvers. Both benchmarks use frame-level mAP. As references, we use the strongest reported video-only methods on each dataset.

\para{HDD.}
Under the official online evaluation protocol, \method achieves 51.3 frame-level mAP, outperforming E2E-LOAD++ (48.4) and SDL (49.1). This demonstrates that the learned representations support online action recognition using only current and past video frames.

\para{ROAD-Waymo.}
\method achieves 63.8 frame-level mAP, exceeding SlowFast (52.7), the strongest video-based baseline reported in the dataset paper, by 11.1 points. Direct label mapping without retraining the classification head also achieves 56.6 mAP, suggesting that the learned representations capture action semantics transferable across different label definitions.

% These are system-level comparisons rather than zero-shot transfer, as \name{} training includes the training videos and converted annotations of both benchmarks.

\subsection{Fine-Grained Action Error Analysis}
\label{sec:error_analysis}

\method still struggles to distinguish lateral actions with similar ego-motion patterns (\cref{fig:confusion}).
Specifically, 34\% of \cls{Right branch} instances are misclassified as \cls{Right lane change}, suggesting difficulty in distinguishing road divergence from lane transitions.
Similarly, 44\% of \cls{Left lane borrowing} instances are confused with \cls{Left nudge}, indicating difficulty in determining whether the vehicle temporarily occupies an adjacent lane.
These errors suggest that fine-grained action recognition requires understanding road topology and lane boundaries beyond ego-motion alone.

\subsection{Natural-Language Video Retrieval Agent}
\label{sec:retrieval_agent}

To demonstrate the practical utility of \method, we build a video retrieval agent using \method's predictions, which provide action probabilities at 0.2-second intervals.
The agent interprets natural-language requests to identify target actions and contextual cues, then searches these temporally indexed predictions to locate matching video intervals.
As shown in \cref{fig:agent_retrieval}, the agent retrieves individual actions, such as \cls{Left lane change} and \cls{Pull-over}, as well as ordered action sequences, such as \cls{Stationary}, \cls{Starting}, and \cls{Left turn}.
This demonstrates the potential of \method as an action perception module for video retrieval.

\section{Conclusion}
\label{sec:conclusion}

We introduced \name, a benchmark that unifies seven public driving datasets into a common fine-grained taxonomy of 16 lateral and 6 longitudinal ego-vehicle actions. By integrating heterogeneous supervision, aligning temporal boundaries, and applying systematic quality control, \name enables standardized training and evaluation of fine-grained ego-action recognition directly from driving video. We further introduced \method, a strong video-based baseline that achieves competitive performance on \name and state-of-the-art results on existing ego-action benchmarks, while demonstrating the practical value of action recognition for targeted driving-video retrieval. Our analysis shows that actions with similar ego motion but different relationships to road topology remain difficult to distinguish, particularly for rare maneuvers. These results establish \name as a foundation for developing models that move beyond coarse vehicle motion toward more detailed understanding and retrieval of driving behavior.

\section*{LLM Usage Statement}
We used commercial large language models to assist with language editing, data processing, and figure creation and refinement. All scientific content, data processing procedures, experimental results, and final figures were reviewed and verified by the authors.

{
    \small
    \bibliographystyle{ieeenat_fullname}
    \bibliography{main}
}
%%%%%%%%% APPENDIX (supplementary material, merged for the arXiv version)
\clearpage
\appendix
\renewcommand{\thesection}{\Alph{section}}
\renewcommand{\thetable}{S\arabic{table}}
\renewcommand{\thefigure}{S\arabic{figure}}
\renewcommand{\theequation}{S\arabic{equation}}
\setcounter{table}{0}\setcounter{figure}{0}\setcounter{equation}{0}
\newcommand{\mainref}[1]{\ref{#1}}
\makeatletter
\def\cvprsect#1{\cvprsection[#1]{\texorpdfstring{\hskip -1em.~}{}#1}}
\def\cvprsubsect#1{\cvprsubsection[#1]{\texorpdfstring{\hskip -1em.~}{}#1}}
\makeatother
\twocolumn[{\centering\Large\bfseries Supplementary Material\par\vspace{1.5em}}]
\noindent Section, figure, table and equation numbers without the prefix S refer to the main paper.

%-------------------------------------------------------------------------
\section{Label System}
\label{sec:labels_details}

The label system generates lateral and longitudinal action labels, ego kinematics, and context flags at 10\,Hz, synchronized with video frames. Source-specific labeling rules are followed by temporal boundary alignment and segment construction.

\subsection{Human Annotations}
\label{sec:labels_human}

\para{HDD.}
We map HDD goal-oriented annotations to turns, U-turns, lane changes, and branches. For merges without directional labels, we infer the direction from turn signals or heading changes, discarding events where neither is available. The \emph{park} annotation maps to \cls{Pull-over}.
Avoidance annotations provide candidates for \cls{Nudge} and \cls{Lane borrowing}, distinguished initially by a peak lateral displacement threshold of 1.5\,m, with direction determined by the displacement sign. All resulting test segments and training/validation nudge and lane-borrowing segments undergo manual review (\cref{sec:quality_details}).
Lane keeping is derived from ego poses (\cref{sec:labels_pose}), while unsupported lateral-motion candidates remain ignored.
Annotation timestamps are aligned with video and CAN signals using the released frame timestamps.

\para{ROAD-Waymo.}
Turn annotations map to the corresponding turns, while \emph{stop} maps to \cls{Stationary}. Overtaking, reversing, and lateral movements without confirmed lane changes are ignored (\cref{sec:quality_details}).
Because ego speed is unavailable, moving segments receive the longitudinal set-valued label \emph{moving}, representing \{\cls{Accelerating}, \cls{Decelerating}, \cls{Cruising}\}.

\subsection{Ego Kinematics}
\label{sec:labels_kin}

We reconstruct speed $v$, acceleration, yaw rate $\omega$, and curvature $\kappa=\omega/|v|$ (set to zero below 1\,m/s), resampled to 10\,Hz.
For Argoverse~2, nuScenes, nuPlan, and PhysicalAI-AV, poses are interpolated and smoothed using a second-order Savitzky--Golay filter over 0.7\,s. Speed is obtained by projecting the smoothed velocity onto the heading, preserving negative values during reversing.
HDD provides vehicle-bus speed and yaw rate.
For BDD-X, we use 1\,Hz GPS speed and course, selecting the gyroscope axis correlated with course rate ($|r|>0.5$), or the course rate otherwise.

\subsection{HD-Map Rules}
\label{sec:labels_map}

\begin{table}[t]
\centering
\caption{\textbf{HMM lane-assignment costs.}
Transition cost $c$ depends on lane connectivity (\cref{eq:hmm}).
The emission cost is $e_t(\ell)=(d/h)^2+2\Delta\theta^2$, where $d$ is lateral distance to the lane center, $h\geq1.5$\,m is half the lane width, and $\Delta\theta$ is heading error. Lanes with $\Delta\theta>60^\circ$ are excluded.}
\label{tab:supp_hmm}
\small
\setlength{\tabcolsep}{4pt}
\begin{tabular}{@{}lc@{}}
\toprule
Lane relation & Cost $c$ \\
\midrule
Same lane & 0 \\
Successor (or its successor) & 0.1 \\
Predecessor & 0.3 \\
Left/right neighbor & 2.0 \\
Neighbor of successor/predecessor & 2.2 \\
Other lane & 6.0 \\
Lane $\leftrightarrow$ off-lane & 3.0 \\
Off-lane $\rightarrow$ off-lane & 0 \\
\bottomrule
\end{tabular}
\end{table}

\para{Lane graphs and assignment.}
Argoverse~2 and nuPlan provide lane polygons and connectivity. For nuScenes, lane connectors represent intersections, and neighboring lanes are inferred from polygon separation ($\leq0.5$\,m), heading difference ($\leq20^\circ$), and lateral offset (2--6\,m).
An off-lane state represents positions outside drivable lane polygons.
We assign each frame to a lane or off-lane state using HMM-based Viterbi decoding (\cref{eq:hmm,tab:supp_hmm}).
Smoothed lane headings (1.1\,s window) provide lateral velocity $v_{\mathrm{lat}}=v\sin(\psi-\psi_{\mathrm{lane}})$ and lane-relative offset $d$.

\para{Action rules.}
A transition between neighboring non-intersection lanes indicates a \cls{Lane change}, extended while $|v_{\mathrm{lat}}|>0.05$\,m/s for up to 6\,s on either side.
It becomes a \cls{Merge} when the original lane ends, with each successor having another predecessor.
A transition to a successor with at least 1\,m lateral displacement and approximately straight motion ($|\omega|\leq0.12$\,rad/s from 2\,s before to 4\,s after) indicates a \cls{Branch} if the original lane splits, or a \cls{Merge} if the target lane joins another.

Intersection-lane sequences are extended while yaw rate exceeds 30\% of its peak. Turns require at least $45^\circ$ heading change; U-turns require at least $150^\circ$ and a turning radius $\leq15$\,m.
The \cls{Lane keeping (curve)} class requires lane curvature above $1/300$\,m$^{-1}$ for at least 2\,s at speeds above 3\,m/s.

A \cls{Pull-over} requires at least 4\,s of stopping near the curb, preceded within 10\,s by at least 0.8\,m of curbward motion and less than $30^\circ$ heading change, away from intersections.
The curb condition requires either an off-lane position or a curb-side lane position at least 0.7\,m closer to the curb than usual and within 2.2\,m of the road edge.
In Argoverse~2, a leading vehicle 2--20\,m ahead and within 1.8\,m of the ego axis indicates queuing rather than pulling over.
The \cls{Pull-out} rule reverses this transition after a stop.
Map-derived nudge and lane-borrowing candidates are excluded from the final labels (\cref{sec:quality_details}).

\subsection{BDD-X Captions}
\label{sec:labels_captions}

We match time-aligned captions to predefined action patterns in priority order: lane changes expressed as ``pulls over into the left/right lane,'' pulling over, pulling out, U-turns, lane changes, and left/right turns.
Common expressions include ``pulls to the curb,'' ``pulls into traffic,'' ``turns around,'' and ``changes/switches/merges into a lane.''
The expression ``around the bend'' is excluded from U-turn matching.
Merging into an adjacent lane is labeled as a lane change, not a merge.
Reversing captions are ignored.
Uncaptioned turns are inferred from poses but discarded when mean speed is below 2\,m/s or peak yaw rate exceeds 1.2\,rad/s to reduce GPS and gyroscope artifacts.

\subsection{Pose-Only Rules}
\label{sec:labels_pose}

Pose-only rules apply to PhysicalAI-AV and lane keeping in HDD and BDD-X.
A turn requires $|\omega|>0.10$\,rad/s (bridging gaps up to 1\,s), at least $45^\circ$ heading change, and a turning radius $\leq40$\,m.
U-turns require at least $150^\circ$, radius $\leq15$\,m, and duration $\leq20$\,s.
The \cls{Lane keeping (curve)} class requires median curvature above $1/300$\,m$^{-1}$ over 2.1\,s, sustained for at least 2\,s above 5\,m/s.

Lane-relative displacements of at least 0.5\,m identify lateral-motion candidates, but cannot reliably distinguish lane changes, obstacle avoidance, or curb maneuvers from road curvature. These candidates are therefore ignored.
A recognition-model-assisted attempt on PhysicalAI-AV achieved only 50--60\% precision in a spot check, below our 80\% criterion, and was excluded from the final labels.

\subsection{Temporal Boundary Alignment}
\label{sec:labels_boundaries}

Map-derived lane changes retain the core interval $\mathcal{T}_{\mathrm{lc}}$ in \cref{eq:boundaries}, with $t_c$ determined by the largest lane-offset jump exceeding 1.5\,m.
Turns derived from poses, maps, or captions retain $\mathcal{T}_{\mathrm{turn}}$ when heading change reaches $30^\circ$ and no more than 20\% of yaw-rate samples are missing.
Core intervals must last at least 0.5\,s.
For other events lasting at least 1.5\,s, including human annotations, we trim 0.5\,s from each boundary.
Trimmed frames remain available as context but are ignored during supervision.

\subsection{Segment Construction}
\label{sec:labels_segments}

\para{Lateral actions.}
Non-lane-keeping maneuvers form segments by joining same-class runs across ignored gaps up to 0.7\,s and adding up to 1\,s of context on either side, without overlapping neighboring maneuvers.
Segments are retained when the target action covers at least 80\% of labeled frames and ignored frames account for no more than 20\%.
Fourteen segments exceed 25.6\,s; their step-token models use the first 25.6\,s.

Lane-keeping intervals of at least 3\,s are split at starts and stops, trimmed around ignored regions with 1\,s margins, and divided into approximately 10\,s segments. Remainders shorter than 3\,s are merged with the preceding segment.

\para{Longitudinal actions.}
The longitudinal rules in \cref{eq:longitudinal} use the first and last $k=\min(10,\lfloor n/5\rfloor)$ frames of an $n$-frame segment.
Segments are discarded if reversing persists for at least 0.5\,s ($v<-0.2$\,m/s), or if speed remains below $v_{\mathrm{s}}$ for at least 1\,s without a \cls{Stationary}, \cls{Starting}, or \cls{Stopping} label.
Lane-keeping segments containing both substantial deceleration and acceleration are split at the speed minimum into pieces of at least 3\,s; otherwise they are discarded.
For non-lane-keeping maneuvers, such speed changes receive the set-valued label \emph{moving}.
ROAD-Waymo uses its stop annotations to distinguish \cls{Stationary}, \cls{Starting}, and \cls{Stopping}; other segments are labeled \emph{moving}.

\para{Scope.}
Night clips are excluded using metadata or mean luminance below 60 in the upper fifth of the image.
Work zones identified in nuScenes metadata and HDD training sessions are excluded with a 4\,s margin. HDD test sessions remain intact under the official protocol.

\para{Dense training labels.}
\method uses 5\,Hz per-step labels before segmentation.
Its training taxonomy additionally includes roundabouts, right lane borrowing, reverse, deceleration to a stop, and set-valued lane keeping for ROAD-Waymo.
Acceleration and deceleration require $|a|>0.5$\,m/s$^2$ for at least 1\,s.

%-------------------------------------------------------------------------
\section{Quality Control}
\label{sec:quality_details}

This section details the consistency checks and manual verification procedures introduced in \cref{sec:quality}.

\begin{table}[t]
\centering
\caption{\textbf{Audit of test labels.}
Per-class label precision and exact 95\% Clopper--Pearson confidence intervals (\%); $n$: checked segments, $k$: correct labels.
Previously reviewed nudges are excluded.}
\label{tab:supp_audit}
\small
\setlength{\tabcolsep}{4pt}
\begin{tabular}{@{}lrrrc@{}}
\toprule
Class & $n$ & $k$ & Prec. & 95\% CI \\
\midrule
\multicolumn{5}{@{}l}{\textit{Lateral}} \\
\quad Lane keeping (straight) & 69 & 69 & 100 & [95, 100] \\
\quad Lane keeping (curve) & 15 & 15 & 100 & [78, 100] \\
\quad Left turn & 28 & 28 & 100 & [88, 100] \\
\quad Right turn & 20 & 20 & 100 & [83, 100] \\
\quad U-turn & 18 & 18 & 100 & [81, 100] \\
\quad Left lane change & 15 & 15 & 100 & [78, 100] \\
\quad Right lane change & 16 & 16 & 100 & [79, 100] \\
\quad Left merge & 18 & 17 & 94 & [73, 100] \\
\quad Right merge & 8 & 8 & 100 & [63, 100] \\
\quad Left branch & 16 & 15 & 94 & [70, 100] \\
\quad Right branch & 15 & 14 & 93 & [68, 100] \\
\quad Left lane borrowing & 12 & 12 & 100 & [74, 100] \\
\quad Pull-over & 5 & 5 & 100 & [48, 100] \\
\quad Pull-out & 6 & 6 & 100 & [54, 100] \\
\quad None (stationary) & 10 & 10 & 100 & [69, 100] \\
\quad All (undecidable: 1) & 271 & 268 & 98.9 & [96.8, 99.8] \\
\midrule
\multicolumn{5}{@{}l}{\textit{Longitudinal}} \\
\quad Stationary & 10 & 10 & 100 & [69, 100] \\
\quad Starting & 21 & 21 & 100 & [84, 100] \\
\quad Stopping & 20 & 20 & 100 & [83, 100] \\
\quad Accelerating & 48 & 48 & 100 & [93, 100] \\
\quad Decelerating & 35 & 35 & 100 & [90, 100] \\
\quad Cruising & 135 & 135 & 100 & [97, 100] \\
\quad \{Accel., Decel., Cruising\} & 3 & 3 & 100 & [29, 100] \\
\quad All (undecidable: 0) & 272 & 272 & 100.0 & [98.7, 100.0] \\
\bottomrule
\end{tabular}
\end{table}

\para{Kinematic consistency.}
We verify labeled actions against heading changes computed from bias-corrected yaw rates (\cref{eq:heading}).
This check removes \nKinRemoved{} inconsistent segments, primarily \cls{Lane keeping (straight)} segments with substantial heading changes (1,481 cases) and turns without sufficient heading changes (149 cases).
Other removals include 55 stationary BDD-X segments with inconsistent heading changes and 9 lane-level or obstacle maneuvers exceeding $45^\circ$.

\para{Image-motion consistency.}
GPS errors in BDD-X can indicate motion while the vehicle is stationary.
We compute the median frame difference over each segment using grayscale images ($160{\times}90$, lower half) separated by 0.5\,s.
The median difference is 15.1 for moving segments and 1.9 for stationary ones.
We remove \nMotionRemoved{} segments labeled as moving, excluding \cls{Starting} and \cls{Stopping}, whose median difference is below 2.

\para{Manual review of obstacle maneuvers.}
HDD avoidance annotations do not specify maneuver direction or whether the vehicle enters an adjacent lane.
We initially infer these properties from lateral displacement and manually review all 27 test nudge candidates, 13 additional ambiguous test segments, and 141 training/validation nudge and lane-borrowing segments.
On the test split, 14 nudges are relabeled as lane borrowings, one changes direction, and four ambiguous segments are removed.
On the training and validation splits, 47 labels are revised and 13 segments are removed.
Reviews use surrounding video without model predictions.

\para{Test label audit.}
We independently audit the final test labels, covering all test U-turns, merges, pull-over and pull-out actions, previously unreviewed lane borrowings, and stratified samples of other lateral and longitudinal classes.
Reviewers inspect randomly ordered video segments with their labels but without model predictions, marking each as correct, incorrect, or undecidable.
As shown in \cref{tab:supp_audit}, 268 of 271 decidable lateral labels (98.9\%) and all 272 longitudinal labels are correct.

%-------------------------------------------------------------------------
\section{Dataset Statistics}
\label{sec:stats_details}

\begin{table*}[t]
\centering
\caption{\textbf{Segment counts by action class, source, and split.}
Set-valued labels are shown in braces.
AV2: Argoverse~2; nuSc.: nuScenes; ROAD-W.: ROAD-Waymo; PAI: PhysicalAI-AV.}
\label{tab:supp_counts}
\small
\setlength{\tabcolsep}{4pt}
\begin{tabular}{@{}l rrrrrrr rrr@{}}
\toprule
 & \multicolumn{7}{c}{Source} & \multicolumn{3}{c}{Split} \\
\cmidrule(lr){2-8}\cmidrule(lr){9-11}
Class & HDD & BDD-X & AV2 & nuSc. & ROAD-W. & nuPlan & PAI & Train & Val & Test \\
\midrule
\multicolumn{11}{@{}l}{\textit{Lateral}} \\
\quad Lane keeping (straight) & 20,769 & 4,822 & 1,363 & 589 & -- & 7,039 & 10,478 & 35,596 & 3,082 & 6,382 \\
\quad Lane keeping (curve) & 263 & 75 & 21 & 43 & -- & 136 & 689 & 1,078 & 49 & 100 \\
\quad Left turn & 1,607 & 305 & 79 & 72 & 83 & 544 & 2,170 & 4,195 & 253 & 412 \\
\quad Right turn & 1,608 & 391 & 129 & 89 & 83 & 378 & 1,839 & 3,821 & 234 & 462 \\
\quad U-turn & 69 & 23 & -- & 1 & -- & 83 & 2,585 & 2,727 & 16 & 18 \\
\quad Left lane change & 570 & 126 & 32 & 7 & -- & 283 & -- & 790 & 105 & 123 \\
\quad Right lane change & 543 & 93 & 31 & 13 & -- & 259 & -- & 732 & 90 & 117 \\
\quad Left merge & 111 & -- & -- & -- & -- & -- & -- & 83 & 11 & 17 \\
\quad Right merge & 46 & -- & -- & -- & -- & -- & -- & 34 & 4 & 8 \\
\quad Left branch & 295 & -- & 1 & -- & -- & -- & -- & 219 & 33 & 44 \\
\quad Right branch & 130 & -- & 6 & -- & -- & -- & -- & 88 & 16 & 32 \\
\quad Left nudge & 64 & -- & -- & -- & -- & -- & -- & 48 & 5 & 11 \\
\quad Right nudge & 10 & -- & -- & -- & -- & -- & -- & 8 & 1 & 1 \\
\quad Left lane borrowing & 82 & -- & -- & -- & -- & -- & -- & 42 & 15 & 25 \\
\quad Pull-over & 43 & 36 & -- & -- & -- & 1 & -- & 68 & 7 & 5 \\
\quad Pull-out & 1 & 22 & -- & -- & -- & 2 & -- & 18 & 1 & 6 \\
\quad None (stationary) & 11,473 & 2,945 & 284 & 224 & 331 & 5,857 & 4,278 & 19,672 & 1,818 & 3,902 \\
\midrule
\multicolumn{11}{@{}l}{\textit{Longitudinal}} \\
\quad Stationary & 11,473 & 2,945 & 284 & 224 & 331 & 5,857 & 4,278 & 19,672 & 1,818 & 3,902 \\
\quad Starting & 210 & 25 & 7 & 3 & 2 & 18 & 84 & 263 & 21 & 65 \\
\quad Stopping & 77 & 24 & 6 & 4 & -- & 4 & 67 & 156 & 6 & 20 \\
\quad Accelerating & 5,608 & 1,319 & 278 & 123 & -- & 1,471 & 3,339 & 9,772 & 791 & 1,575 \\
\quad Decelerating & 5,836 & 1,456 & 388 & 144 & -- & 1,924 & 3,238 & 10,312 & 876 & 1,798 \\
\quad Cruising & 14,399 & 3,050 & 970 & 537 & -- & 5,307 & 10,922 & 28,741 & 2,203 & 4,241 \\
\quad \{Accel., Decel., Cruising\} & 81 & 19 & 13 & 3 & 164 & 1 & 111 & 303 & 25 & 64 \\
\midrule
All segments & 37,684 & 8,838 & 1,946 & 1,038 & 497 & 14,582 & 22,039 & 69,219 & 5,740 & 11,665 \\
\bottomrule
\end{tabular}
\end{table*}

\para{Video preprocessing.}
All videos are standardized to 10\,fps and $512{\times}288$ resolution (16:9), with frames sampled at the timestamps of their corresponding action labels.
Argoverse~2's portrait images are cropped to a 16:9 region starting at 30\% of the image height, while ROAD-Waymo's 3:2 images are cropped around the horizon.
The other sources use their front-camera views, including the $120^\circ$ camera of PhysicalAI-AV.

\para{Class distribution.}
\Cref{tab:supp_counts} reports segment counts by action class, source, and split.
HDD contributes all merges, nudges, and lane borrowings and most branches; BDD-X provides most curb maneuvers, while PhysicalAI-AV contributes primarily lane keeping and turns.
The longitudinal set-valued label \{\cls{Accelerating}, \cls{Decelerating}, \cls{Cruising}\} occurs predominantly in turns and U-turns (388 of 392 segments), reflecting either speed changes within a maneuver or unavailable speed measurements.

\para{Geographic coverage.}
The recordings span \nCountries{} countries, with \hUS{}\,h from the United States and the remainder from Europe and Singapore.
Left-hand traffic is represented only by approximately 0.8\,h of nuScenes recordings from Singapore.

%-------------------------------------------------------------------------
\section{Evaluation Protocol}
\label{sec:protocol_details}

\para{Evaluated classes.}
Lateral actions are evaluated on \nTestMoving{} moving test segments, while longitudinal actions are evaluated on all \nTest{} segments.
The headline mAP (\cref{eq:map}) covers eight lateral classes (\cls{Lane keeping (curve)}, \cls{Left/right turn}, \cls{Left/right lane change}, \cls{Left/right branch}, and \cls{Left lane borrowing}) and five longitudinal classes (\cls{Starting}, \cls{Stopping}, \cls{Accelerating}, \cls{Decelerating}, and \cls{Cruising}).
The background classes \cls{Lane keeping (straight)} and \cls{Stationary} contribute negatives but are excluded from mAP.
The seven remaining lateral classes have fewer than 20 test examples and are evaluated using top-1 accuracy ($k/n$) with exact 95\% Clopper--Pearson confidence intervals. Their mean accuracy is reported in the main results.

\para{Set-valued labels.}
For class $c$, a segment labeled with set $S$ is positive if $S=\{c\}$, negative if $c\notin S$, and excluded if $c\in S$ and $|S|>1$.
Classwise AP is computed from $\hat p_c$ using \texttt{average\_precision\_score} in scikit-learn.

\para{Auxiliary metrics.}
Macro-F1 is computed from the top-scoring prediction across the evaluated classes and background class of each axis.
The false-alarm rate is the proportion of \cls{Lane keeping (straight)} segments predicted as a lateral maneuver other than \cls{Lane keeping (curve)}.
For each action family $F$, family AP is computed by summing member-class probabilities and treating segments labeled within $F$ as positives and those outside $F$ as negatives; ambiguous sets intersecting $F$ are excluded.
The families are turn (\cls{Left/right turn}, \cls{U-turn}), lane-level (\cls{Left/right lane change}, \cls{Left/right merge}, \cls{Left/right branch}), obstacle (\cls{Left/right nudge}, \cls{Left lane borrowing}), and curb (\cls{Pull-over}, \cls{Pull-out}).

\para{Statistical reliability.}
Results are averaged over three random seeds and reported as mean $\pm$ standard deviation.
We estimate 95\% confidence intervals using 1,000 source-stratified cluster bootstrap resamples of complete recordings.
For each resample, we compute seed-averaged mAP while keeping the evaluated classes fixed, and use the 2.5th and 97.5th percentiles as interval bounds.
Paired comparisons use identical resamples across models and compute intervals from their per-resample performance differences.

%-------------------------------------------------------------------------
\section{Implementation Details}
\label{sec:impl_details}

\begin{table}[t]
\centering
\caption{\textbf{Training configurations of \method.}
Stage~1 consists of a frozen-encoder warm-up followed by full fine-tuning; Stage~2 trains the segment head with the adapted encoder frozen.}
\label{tab:supp_hparams}
\small
\setlength{\tabcolsep}{2pt}
\begin{tabular}{@{}lll@{}}
\toprule
Configuration & Stage 1 & Stage 2 \\
\midrule
Input & 25.6\,s (128 steps) & segment ($\leq128$ steps) \\
Labels & per-step (0.2\,s) & segment-level \\
Trainable & heads $\rightarrow$ full model & segment head \\
Steps & 6,000 + 10,000 & 6,000 \\
Batch size & 32 / 8 & 64 \\
Learning rate & $3{\times}10^{-4}$ / layer-wise & $3{\times}10^{-4}$ \\
Optimizer & \multicolumn{2}{l}{AdamW, weight decay 0.05, gradient clipping 1.0} \\
Schedule & \multicolumn{2}{l}{one-cycle, 5\% warm-up, cosine decay} \\
\bottomrule
\end{tabular}
\end{table}

\para{Input and encoder.}
Videos are stored at 10\,fps and $512{\times}288$ resolution, resized to $224{\times}448$ without cropping, and normalized using ImageNet statistics.
We use V-JEPA~2 ViT-L (\texttt{vjepa2-vitl-fpc64-256}), which independently encodes non-overlapping 16-frame windows into $8{\times}14{\times}28$ tokens.
Each temporal token corresponds to 0.2\,s.
For RoPE interpolation, column $w$ is assigned position $15w/27$, while spatial row and temporal positions remain unchanged.

\para{Spatial readout and temporal model.}
At each step, global averaging and $4{\times}8$ spatial pooling of the final-layer tokens yield 33 features.
After layer normalization and projection to 384 dimensions, learned spatial embeddings and a six-head attention query produce the step representation (\cref{eq:readout}).
The temporal model uses learned positional embeddings and four pre-norm transformer layers (width 384, six attention heads, feed-forward dimension 1536, dropout 0.1).

\para{Stage 1: Dense supervision.}
Prediction heads supervise lateral and longitudinal actions, HDD goal-oriented classes, and four ego-kinematic quantities.
Predicted kinematics also condition the action heads through a 4--64--64 MLP.
Kinematic targets are normalized by 10\,m/s, 2\,m/s$^2$, 0.3\,rad/s, and 0.05\,m$^{-1}$, clipped to $\pm5$, and masked when unavailable.
HDD goal supervision uses steps within 0.2\,s of official 3\,fps annotations.

Training samples are drawn from 17,438 training clips (206\,h), with probability proportional to clip duration divided by the square root of source size.
After the frozen-encoder warm-up, full fine-tuning uses a learning rate of $5{\times}10^{-5}\cdot0.8^{23-i}$ for encoder block $i\in\{0,\ldots,23\}$, $5{\times}10^{-5}\cdot0.8^{24}$ for patch embedding, $5{\times}10^{-5}$ for the final layer normalization, and $10^{-4}$ for the prediction heads, with gradient checkpointing.

\para{Stage 2: Segment-level recognition.}
We freeze the adapted encoder and cache its step tokens in fp16.
The segment head reuses the spatial readout and temporal transformer, followed by attention pooling and separate lateral and longitudinal classifiers.
It predicts 17 lateral logits (16 actions plus stationary) and seven longitudinal logits (six actions plus reverse), using the set-valued loss on both axes.
Three runs vary the head initialization and training samples.

\begin{table}[t]
\centering
\caption{\textbf{Configurations of existing video models.}
Public checkpoint sources, pre-training datasets, input dimensions, and output features.
PTV: PyTorchVideo; HF: Hugging Face.}
\label{tab:supp_models}
\small
\setlength{\tabcolsep}{2pt}
\begin{tabular}{@{}lllc l@{}}
\toprule
Model & Checkpoint & Pre-train & Input & Feature \\
\midrule
TSN-R50 & timm & IN-1K & $8\times224^2$ & frame mean \\
I3D-R50 & PTV & K400 & $8\times224^2$ & pooled \\
X3D-M & PTV & K400 & $16\times224^2$ & pooled \\
SlowFast-R50 & PTV & K400 & $32\times256^2$ & pooled \\
TimeSformer-B & HF & K400 & $8\times224^2$ & class token \\
VideoMAE-B & HF & K400 & $16\times224^2$ & token mean \\
\bottomrule
\end{tabular}
\end{table}

\para{Existing models.}
Each backbone is followed by layer normalization and separate lateral and longitudinal classifiers.
Frames are uniformly sampled within temporal bins during training and at bin centers during evaluation, then resized to square inputs without cropping or augmentation.
PTV models use normalization with mean 0.45 and standard deviation 0.225; other models use ImageNet statistics.

All models are trained for 4,000 steps with batch size 32 using AdamW (learning rates $10^{-4}$ for the backbone and $10^{-3}$ for the heads, weight decay 0.05), a one-cycle schedule with 5\% warm-up, bf16 autocast, and gradient clipping at 1.0.
They share the set-valued loss and class-balanced sampling.

The frozen V-JEPA~2 baselines use the same Stage~2 head with either a global token and $2{\times}4$ pooled grid or a global token and $4{\times}8$ grid.

\para{Model selection.}
The validation split was introduced after the main architectural decisions.
RoPE interpolation was selected using HDD test sessions, while spatial readout, temporal context, and full fine-tuning were compared on an earlier test pool that included ROAD-Waymo validation videos.
These choices may favor \method on the test split.
The existing models follow a common training recipe without model-specific tuning, apart from the schedule check above.
Ablations are therefore also reported on the validation split (\cref{tab:ablation}).

%-------------------------------------------------------------------------
\section{Additional results}
\label{sec:results_details}

\begin{table}[t]
\centering
\caption{\textbf{Paired 95\% confidence intervals} for the mAP difference between \method and each baseline on the test split, estimated using 1,000 cluster bootstrap resamples.}
\label{tab:supp_paired}
\small
\setlength{\tabcolsep}{4pt}
\begin{tabular}{@{}lcc@{}}
\toprule
Method & Lateral & Longitudinal \\
\midrule
TSN-R50~\cite{wang2016tsn} & [28.0, 38.1] & [46.8, 56.1] \\
I3D-R50~\cite{carreira2017i3d} & [6.0, 13.1] & [14.3, 21.4] \\
X3D-M~\cite{feichtenhofer2020x3d} & [9.9, 17.9] & [12.1, 19.2] \\
SlowFast-R50~\cite{feichtenhofer2019slowfast} & [2.8, 8.2] & [4.3, 8.9] \\
TimeSformer-B~\cite{bertasius2021timesformer} & [6.1, 13.8] & [32.2, 40.2] \\
VideoMAE-B~\cite{tong2022videomae} & [$-$1.9, 4.4] & [2.6, 7.5] \\
V-JEPA~2-L, frozen, pooled & [3.7, 9.2] & [0.5, 4.0] \\
V-JEPA~2-L, frozen, grid & [2.4, 8.0] & [0.9, 4.6] \\
\bottomrule
\end{tabular}
\end{table}

\para{Paired comparisons.}
\Cref{tab:supp_paired} reports paired confidence intervals for the mAP differences between \method and the baselines.
All intervals exclude zero except the lateral comparison with VideoMAE, indicating comparable lateral performance but higher longitudinal mAP for \method.

\begin{table*}[t]
\centering

\caption{\textbf{Performance by data source} (test split, mean over three seeds). HDD mAP follows the headline lateral protocol restricted to HDD. Mean recall averages lateral action classes with at least 20 test segments per source (HDD: 9; BDD-X: 3; AV2: 2; nuSc.: 1; ROAD-W.: 1).}

\label{tab:supp_source}
\small
\setlength{\tabcolsep}{3pt}
\begin{tabular}{@{}l c ccccc@{}}
\toprule
 & mAP & \multicolumn{5}{c}{Mean recall} \\
\cmidrule(lr){2-2}\cmidrule(lr){3-7}
Method & HDD & HDD & BDD-X & AV2 & nuSc. & ROAD-W. \\
\midrule
TSN & 54.6 & 58.7 & 64.6 & 88.6 & 84.1 & 80.6 \\
I3D & 76.8 & 74.2 & 90.0 & 98.0 & 90.6 & 90.3 \\
X3D & 72.2 & 71.2 & 89.6 & 98.1 & 90.2 & 90.3 \\
SlowFast & 80.3 & 74.0 & 93.6 & 98.9 & 92.0 & 93.1 \\
TimeSformer & 76.1 & 71.9 & 88.9 & 98.8 & 89.9 & 93.1 \\
VideoMAE & 83.5 & 76.7 & 95.4 & \textbf{99.3} & 95.3 & 91.7 \\
V-JEPA 2 frozen, pooled & 79.8 & 76.5 & 95.5 & 98.9 & 96.4 & \textbf{95.8} \\
V-JEPA 2 frozen, grid & 80.7 & 75.2 & \textbf{96.1} & 99.0 & 96.0 & 91.7 \\
\rowcolor{oursrow}\method & \textbf{85.0} & \textbf{80.6} & 95.9 & 99.1 & \textbf{97.5} & 93.1 \\
\bottomrule
\end{tabular}
\end{table*}

\para{Performance across sources.}
\Cref{tab:supp_source} shows that most non-HDD sources achieve high recall, as their evaluated classes primarily involve lane keeping and turns.
HDD is more challenging because it contains lane-level and obstacle-avoidance actions, accounting for much of the performance variation across models.

\section{Evaluation on Existing Benchmarks}
\label{sec:transfer_details}

\para{Training data and comparison scope.}
The training split of \name includes the official training videos of HDD and ROAD-Waymo, with annotations converted to our action taxonomy (\cref{sec:labels_human}).
Stage~1 additionally uses HDD goal-oriented labels.
Neither benchmark's validation nor test videos are used for training.
For benchmark-specific evaluation, we freeze the adapted encoder and train a task-specific head on the respective training split.
These are system-level comparisons rather than zero-shot transfer: unlike the published methods trained on the target dataset alone, our encoder is jointly trained on the target training data and five additional datasets.

\para{HDD.}
We follow the official online protocol with 11 goal-oriented action classes.
Each prediction uses only the preceding 16 frames, and the Stage~1 head is retrained with causal attention on the 100 official training sessions (36 test sessions).
Training uses action labels and vehicle-bus kinematics, with the latter unavailable as model inputs at inference.
The head is trained for 8,000 steps using AdamW (learning rate $3{\times}10^{-4}$, batch size 32).

At inference, 12.8\,s windows ($L=64$) slide with stride $L/2$, providing 6.4--12.8\,s of history per prediction.
We evaluate all 272,709 official test frames at 3\,fps, averaging AP over the 11 action classes without background.
The three runs achieve 51.2, 51.6, and 51.0 frame mAP.
Extending the window to $L=128$ yields 47.1 mAP, while bidirectional temporal modeling achieves 54.3 mAP in the offline setting.

\para{ROAD-Waymo.}
We follow the official validation protocol for six AV actions (move, stop, turn left/right, move left/right), evaluated using frame-level AP over annotated frames at 10\,Hz.
The task-specific head consists of layer normalization, a 256-dimensional projection, learned slot embeddings, attention readout, and two transformer layers.
It predicts six sigmoid outputs and is trained with binary cross-entropy for 3,000 steps using AdamW (learning rate $3{\times}10^{-4}$, batch size 16, 25.6\,s windows).
The three runs obtain 64.3, 63.1, and 63.9 mAP.

Without retraining a head, direct mapping of Stage~1 predictions achieves 56.6 mAP.
The mapping assigns \cls{Stationary} to \emph{stop}, non-stationary and non-reverse to \emph{move}, \cls{Left turn} and \cls{U-turn} to \emph{turn left}, \cls{Right turn} to \emph{turn right}, and directional lane-level or obstacle actions to \emph{move left/right}.

%-------------------------------------------------------------------------
\section{Natural-Language Video Retrieval}
\label{sec:agent}

\begin{figure*}[t]
\centering
\includegraphics[width=\textwidth]{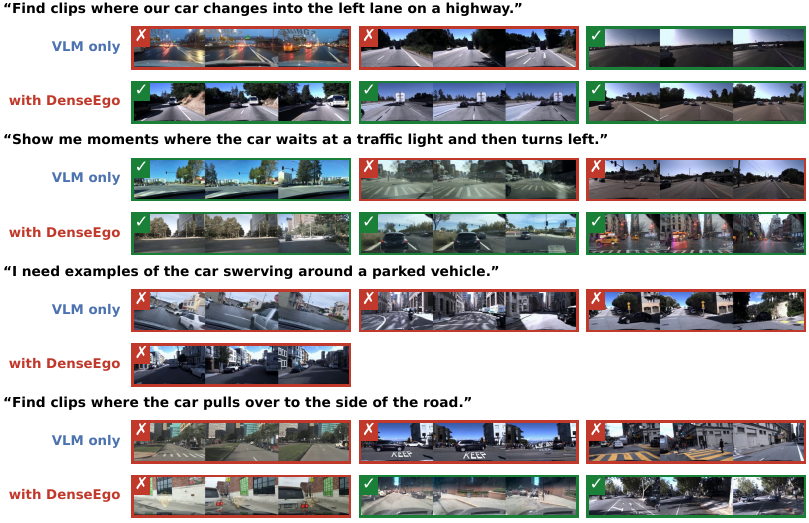}
\caption{\textbf{Fine-grained ego-action retrieval from natural-language queries.}
Three queries cover single actions and an ordered action sequence. For each query, the top three results from VLM-only and \method-assisted retrieval are compared, with retrieval scores shown below each clip. Green indicates that the retrieved span matches the requested actions according to benchmark labels; red indicates a mismatch.}
\label{fig:agent}
\end{figure*}

\para{Retrieval pipeline.}
A local Qwen3-VL-8B model parses each natural-language query into ordered ego-action targets, scene constraints, and a binary verification question.
The same model indexes scene context from one frame every 10\,s across 33.6\,h of test video, using a fixed vocabulary of road types, intersections, weather, lighting, traffic, and visible objects.
For action retrieval, \method's Stage~1 predictions are converted into events by grouping consecutive steps whose smoothed class probability exceeds 0.5.
Candidate events are matched in the requested temporal order, filtered by scene constraints, and ranked by their mean action probabilities. Retrieval takes approximately one second per query.

\para{VLM-only baseline.}
Without \method, the same agent filters candidate 8\,s windows by scene constraints and uses Qwen3-VL-8B to score each window according to its probability of answering ``yes'' to the verification question.
Up to 800 candidate windows are evaluated per query, taking approximately seven minutes.

\para{Qualitative results.}
As shown in \cref{fig:agent}, the VLM-only baseline frequently assigns similarly high scores to both matching and non-matching clips, limiting its ability to distinguish fine-grained actions.
The action index provides more discriminative retrieval results, including temporally ordered actions.

\para{Limitations.}
The Stage~1 head sometimes predicts upcoming turns while the vehicle is stationary, when lateral labels are ignored during training. We therefore exclude stationary intervals from lateral-action events.
Retrieval also depends on accurate query parsing and recognition of rare actions. For example, an obstacle-avoidance query was incorrectly parsed as requiring both a nudge and a lane borrowing, yielding an incorrect result.

%-------------------------------------------------------------------------
\section{Release and Responsible Use}
\label{sec:release}

\para{Release and data access.}
We release the code for generating \name labels and segments from the seven source datasets, including manual corrections (\cref{sec:quality_details}) and evaluation scripts.
No original videos, images, sensor measurements, maps, or annotations are redistributed. Users must obtain the source datasets under their respective licenses and reconstruct the benchmark locally.

\para{Data usage and privacy.}
PhysicalAI-AV is used exclusively for training and excluded from validation and testing, in accordance with its license.
Source videos may contain identifiable road users or license plates and remain subject to their providers' privacy policies.
Our annotations describe only ego-vehicle actions and introduce no additional information about other road users.

\end{document}